\def\year{2020}\relax
\documentclass[letterpaper]{article} 
\usepackage{aaai20}  
\usepackage{times}  
\usepackage{helvet} 
\usepackage{courier}  
\usepackage[hyphens]{url}  
\usepackage{graphicx} 
\usepackage{multirow}
\usepackage{array}
\usepackage{booktabs}
\usepackage{hhline}
\usepackage{tabularx}
\usepackage{colortbl}
\usepackage[table,xcdraw]{xcolor}
\usepackage{graphicx}  
\definecolor{amber}{rgb}{1.0, 0.75, 0.0}
\definecolor{mi_yellow}{rgb}{1.0, 0.84, 0.4}
\definecolor{mi_pink}{rgb}{0.92, 0.54, 0.83}

\title{Algorithmic Discrimination: Formulation and Exploration \\ in Deep Learning-based Face Biometrics}
\author{Ignacio Serna,\textsuperscript{\rm 1} Aythami Morales,\textsuperscript{\rm 1} Julian Fierrez,\textsuperscript{\rm 1} Manuel Cebrian,\textsuperscript{\rm 2} Nick Obradovich,\textsuperscript{\rm 2} Iyad Rahwan\textsuperscript{\rm 2}  \\ 
\textsuperscript{\rm 1}Universidad Autonoma de Madrid, Madrid, Spain\\
\textsuperscript{\rm 2}Max Planck Institute for Human Development, Berlin, Germany\\
\textsuperscript{\rm 1}\{ignacio.serna, aythami.morales, julian.fierrez\}@uam.es\\
\textsuperscript{\rm 2}\{cebrian, obradovich, sekrahwan\}@mpib-berlin.mpg.de\\
}

\begin{document}

\maketitle

\begin{abstract}
The most popular face recognition benchmarks assume a distribution of subjects without much attention to their demographic attributes. In this work, we perform a comprehensive discrimination-aware experimentation of deep learning-based face recognition. The main aim of this study is focused on a better understanding of the feature space generated by deep models, and the performance achieved over different demographic groups. We also propose a general formulation of algorithmic discrimination with application to face biometrics. The experiments are conducted over the new DiveFace database composed of 24K identities from six different demographic groups\footnote{Available at GitHub: https://github.com/BiDAlab/DiveFace}. Two popular face recognition models are considered in the experimental framework: ResNet-50 and VGG-Face.  We experimentally show that demographic groups highly represented in popular face databases have led to popular pre-trained deep face models presenting strong algorithmic discrimination. That discrimination can be observed both qualitatively at the feature space of the deep models and quantitatively in large performance differences when applying those models in different demographic groups, e.g. for face biometrics.
\end{abstract}

\section{Introduction}
Face recognition algorithms are good examples of recent advances in Artificial Intelligence (AI). The performance of automatic face recognition has been boosted during the last decade, achieving very competitive accuracies in the most challenging scenarios \cite{1}. These improvements have been possible due to improved machine learning approaches (e.g., deep learning), powerful computation (e.g., GPUs), and larger databases (e.g., at scale of millions of images). However, the recognition accuracy is not the only aspect to consider when designing biometric systems. Algorithms have an increasingly important role in the decision-making of several processes involving humans. These decisions have therefore increasing effects in our lives. Thus, there is currently a growing need for studying AI behavior to better understand its impact in our society \cite{31}.

Face recognition systems are especially sensitive due to the personal information present in face images (e.g., identity, gender, ethnicity, and age). Previous works suggested that face recognition accuracy is affected by demographic covariates. In \cite{6,5}, authors demonstrated that the performance of commercial face recognition systems varies according to demographic attributes. In \cite{7,8}, the authors evaluated how covariates affect the performance of face recognition systems based on deep neural network models. Among the different covariates, the skin color is repetitively remarked as a factor with high impact in the performance \cite{6,7}. However, ethnic face attributes are beyond skin color. The shape and size of facial features are partially defined by the ancestry origin. These differences can be used to accurate classify subjects according to their ancestry origin \cite{8}. 

\begin{figure*}[t]
\centering
\includegraphics{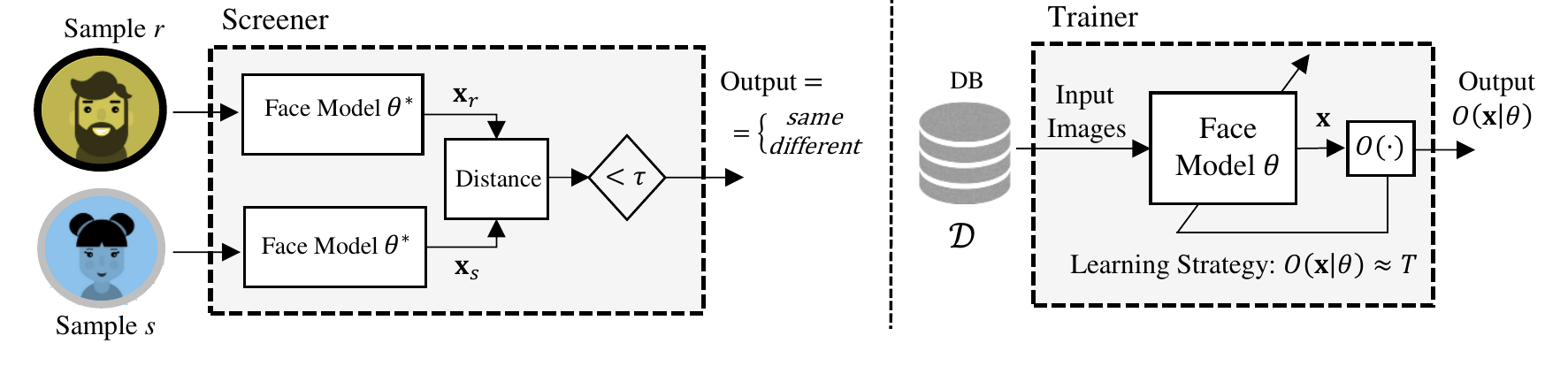} 
\caption{Face recognition block diagrams. The screener is an algorithm that given two face images decides if they belong to the same person. The trainer is an algorithm that generates the best data representation for the screener.}
\label{Figure1}
\end{figure*}

The number of published works pointing out the biases in the results of face detection \cite{3} and recognition algorithms is large \cite{5,8,4,6,7,32}. Yet, only a limited number of works analyze how biases affect the learning process of these algorithms. The aim of this work is to analyze face recognition models using a discrimination-aware perspective. Previous studies have demonstrated that ethnicity and gender affect the performance of face recognition models \cite{gong2019debface}. However, there is a lack of understanding regarding how this demographic information affects the model beyond the performance. The main contributions of this work are: 
\begin{quote}
\begin{itemize}
\item A general formulation of algorithmic discrimination for machine learning tasks. In this work, we apply this formulation in the context of face recognition.
\item Discrimination-aware performance analysis based on a new dataset \cite{9}, with 24K identities equally distributed between six demographic groups.
\item Study of the effects of gender and ethnicity in the feature representation of deep models.
\item Analysis of the demographic diversity present in some of the most popular face databases.
\end{itemize}
\end{quote}

The rest of the paper is structured as follows: Section 2 presents our general formulation of algorithmic discrimination. Section 3 analyzes some of the most popular face recognition architectures and the experimental protocol followed in this work. Section 4 evaluates the causes and effects of biased learning in face recognition algorithms.  Finally, Section 5 summarizes the main conclusions.

\section{Formulation of Algorithmic Discrimination}
Discrimination is defined by the Cambridge Dictionary as treating a person or particular group of people differently, especially in a worse way than the way in which you treat other people, because of their skin color, sex, sexuality, etc.

For the purpose of studying discrimination in artificial intelligence at large, we now formulate mathematically algorithmic discrimination based on the previous dictionary definition. Even though similar ideas as the ones embedded in our formulation can be found elsewhere \cite{23,22}, we didn't find this kind of formulation in related works. We hope that formalizing these concepts can be beneficial to foster further research and discussion in this hot topic.

Let’s begin with notation and preliminary definitions. Assume $\textbf{x}_s^i$ is a learned representation of individual $i$ (out of $I$ different individuals) corresponding to an input sample $s$ (out of $S$ samples) of that particular subject. That representation $\textbf{x}$ is assumed to be useful for task $T$, e.g., face authentication or emotion recognition. That representation $\textbf{x}$ is learned using an artificial intelligence approach with parameters $\theta$. We also assume that there is a goodness criterion $G$ on that task maximizing some performance real-valued function $f$ in a given dataset $\mathcal{D}$ (collection of multiple samples) in the form:

\begin{equation}
\label{eqn:goodnes_criterion}
    \textit{G}(\mathcal{D}) = \max_{\theta}\textit{f}(\mathcal{D},\theta)
\end{equation}

The most popular form of the previous expression minimizes a loss function $\mathcal{L}$ over a set of training samples $\mathcal{D}$ in the form:

\begin{equation}
\label{eqn:learning_strategy}
    \theta^*=\arg\min_{\theta}{\sum_{\textbf{x}_s^i\in \mathcal{D}}\mathcal{L}(\textit{O}(\textbf{x}_s^i|\theta),T^i_s)} 
\end{equation}
where \textit{O} is the output of the learning algorithm that we seek to bring closer to the target function (or groundtruth) \textit{T} defined by the task at hand. On the other hand, the \textit{I} individuals can be classified according to \textit{D} demographic criteria $\textit{C}_d$, with $d = 1,..., D$, which can be the source for discrimination, e.g., $\textit{C}_1 = \textit{Gender} = \{\textit{Male, Female}\}$ (demographic criterion $1 = \textit{Gender}$ has two classes in this example). The particular class $k=1,...,K$ for a given demographic criterion $d$ and a given sample is noted as $\textit{C}_d (\textbf{x}_s^i)$, e.g., $\textit{C}_1 (\textbf{x}_s^i)=\textit{Male}$. We assume that all classes are well represented in dataset $\mathcal{D}$, i.e., the number of samples for each class in all criteria in $\mathcal{D}$ is significant. $\mathcal{D}_d^k \in \mathcal{D}$ represents all the samples corresponding to class $k$ of demographic criterion $d$.

Finally, \textbf{our definition of algorithmic discrimination}: an algorithm discriminates the group of people represented with class $k$ (e.g., \textit{Female}) when performing the task \textit{T} (e.g., face verification, or emotion recognition), if the goodness \textit{G} in that task when considering the full set of data $\mathcal{D}$ (including multiple samples from multiple individuals), is significantly larger than the goodness $\textit{G}(\mathcal{D}_d^k)$ in the subset of data corresponding to class $k$ of the demographic criterion $d$.

The representation $\textbf{x}$ and the model parameters $\theta$ will typically be real-valued vectors, but they can be any set of features combining real and discrete values. Note that the previous formulation can be easily extended to the case of varying number of samples $S_i$  for different subjects, which is a usual case; or to classes \textit{K} that are not disjoint. Note also that the previous formulation is based on average performances over groups of individuals. Different performance across specific individuals is usual in many artificial intelligence tasks due to diverse reasons, e.g., specific users who were not sensed properly \cite{24}, even for algorithms that on average may perform similarly for the different classes that can be the source of discrimination.

\section{Face Recognition Algorithms}
A face recognition algorithm, as other machine learning systems, can be divided into two different algorithms: screener and trainer. Both algorithms are used for a different aim and therefore should be studied with a different perspective \cite{33}.

The screener (see Fig. \ref{Figure1}) is an algorithm that given two face images generates an output associated to the probability that they belong to the same person. This probability is obtained comparing the two learned representations obtained from a face model defined by the parameters $\theta$. These parameters are trained previously based on a training dataset $\mathcal{D}$ and the goodness criterion \textit{G} (see Fig. \ref{Figure1}). If trained properly, the output of the trainer would be a model with parameters $\theta^*$ capable of representing the input data (e.g., face images) in a highly discriminant feature space $\textbf{x}$.

The most popular architecture used to model face attributes is the Convolutional Neural Network (CNN). This type of network has drastically reduced the error rates of face recognition algorithms in the last decade \cite{28} by learning highly discriminative features from large-scale databases. In our experiments we consider two popular face recognition pre-trained models: VGG-Face and ResNet-50. These models have been tested on competitive evaluations and public benchmarks \cite{13,12}. 

VGG-Face is a model based on the VGG-Very-Deep-16 CNN architecture trained on the VGGFace dataset \cite{13}. ResNet-50 is a CNN model with 50 layers and 41M parameters initially proposed for general purpose image recognition tasks \cite{29}. The main difference between ResNet architecture and traditional convolutional neural networks is the inclusion of residual connections to allow information to skip layers and improve gradient flow. 

Before applying the face models, we cropped the face images using the algorithm proposed in \cite{26}. The pre-trained models are used as embedding extractor where $\textbf{x}$ is a $l_2$-normalised learned representation of a face image. The similarity between two face descriptors $\textbf{x}_r$ and $\textbf{x}_s$ is calculated as the Euclidean distance $||\textbf{x}_r-\textbf{x}_s||$. Two faces are assigned to the same identity if their distance is smaller than a threshold $\tau$. The recognition accuracy is obtained by comparing distances between positive matches (i.e., $\textbf{x}_r$  and $\textbf{x}_s$ belong to the same person) and negative matches (i.e., $\textbf{x}_r$  and $\textbf{x}_s$ belong to different persons).

The two face models considered in our experiments were trained with the VGGFace2 dataset according to the details provided in \cite{12}. As we will show in Section \ref{Bias in face databases}, databases used to train these two models are highly biased. Therefore, it is expected that the recognition models trained with this dataset present algorithmic discrimination.

\subsection{Experimental protocol}

Labeled Faces in the Wild (LFW) is a database for research on unconstrained face recognition \cite{20}. The database contains more than 13K images of faces collected from the web. In this study we consider the aligned images from the test set provided with view 1 and its associated evaluation protocol. This database is composed by images acquired in the wild, with large pose variations, and varying face expressions, image quality, illuminations, and background clutter among other variations. The performance achieved by the VGG-Face and ResNet-50 models for the LFW database is $4.1\%$ and $1.7\%$ Equal Error Rate respectively. These performances serve as a baseline for the  models and the rest of experiments. We can observe the superior performance of the ResNet-50 model, with a performance ca. 3 times better than the VGG-Face model. 

The experiments with DiveFace will be carried out following a cross-validation methodology using three images for each of the 4K identities from each of the six classes available in DiveFace (72K face images in total). This results in 72K genuine comparisons and near 3M impostor comparisons.

\subsection{DiveFace database: an annotation dataset for face recognition trained on diversity}

DiveFace was generated using the Megaface MF2 training dataset \cite{11}. MF2 is part of the publicly available Megaface dataset with 4.7 million faces from 672K identities and it includes their respective bounding boxes. All images in the Megaface dataset were obtained from Flickr Yahoo's dataset \cite{27}. 

DiveFace contains annotations equally distributed among six classes related to gender and ethnicity (see Fig. \ref{Figure4} for example images). Gender and ethnicity have been annotated following a semi-automatic process. There are 24K identities (4K for class). The average number of images per identity is 5.5 with a minimum number of 3 for a total number of images greater than 120K. Users are grouped according to their gender (male or female) and three categories related with ethnic physical characteristics:

\begin{quote}
\begin{itemize}
\item \textbf{Group 1}: people with ancestral origins in Europe, North-America, and Latin-America (with European origin).
\item \textbf{Group 2}: people with ancestral origins in Sub-Saharan Africa, India, Bangladesh, Bhutan, among others.
\item \textbf{Group 3}: people with ancestral origin in Japan, China, Korea, and other countries in that region.
\end{itemize}
\end{quote}

We are aware of the limitations of grouping all human ethnic origins into only three categories. According to studies, there are more than 5K ethnic groups in the world. We categorized according to only three groups in order to maximize differences among classes. Automatic classification algorithms based on these three categories show performances of up to 98\% accuracy \cite{9}.

\section{Causes and Effects of Biased Learning in Face Recognition Algorithms}
\subsection{Performance of face recognition: role of demographic information}

\begin{table*}[ht]
\centering
\caption{Performance (False Match Rate in \% @ False Non-Match Rate = 0.1\%) of Face Recognition Models on the DiveFace dataset. We show in brackets the relative error growth rates with respect to the best class (\textit{Group 1 Male}).}\smallskip
\begin{tabular}{@{}>{\centering}p{2cm}>{\centering}p{2.2cm}>{\centering}p{2.2cm}
                 >{\centering}p{2.2cm}>{\centering}p{2.2cm}>{\centering}p{2.2cm}c@{}}
        \cmidrule(l){1-7}
   \multirow{2}{*}{\textbf{Model}} & \multicolumn{2}{c}{\textcolor{mi_yellow}{\textbf{Group 1}}}
                                        & \multicolumn{2}{c}{\textcolor{mi_pink}{\textbf{Group 2}}}
                                            & \multicolumn{2}{c}{\textcolor{blue}{\textbf{Group 3}}}\\
        \cmidrule(lr){2-3}\cmidrule(lr){4-5}\cmidrule(l){6-7}
    & \textbf{Male} & \textcolor{gray}{\textbf{Female}} & \textbf{Male} 
                & \textcolor{gray}{\textbf{Female}} & \textbf{Male} & \textcolor{gray}{\textbf{Female}} \\ 
        \cmidrule(r){1-1}
        \cmidrule(lr){2-2}\cmidrule(lr){3-3}
        \cmidrule(lr){4-4}\cmidrule(lr){5-5}
        \cmidrule(lr){6-6}\cmidrule(l){7-7}
    VGG-Face & 7.99  & 9.38 ($\uparrow$17\%) & 12.03 ($\uparrow$50\%) & 13.95 ($\uparrow$76\%) & 18.43 ($\uparrow$131\%) & 23.66 ($\uparrow$196\%) \\
    ResNet-50 & 1.60  & 1.96 ($\uparrow$22\%) & 2.15 ($\uparrow$34\%) & 3.61 ($\uparrow$126\%) & 3.25 ($\uparrow$103\%) & 5.07 ($\uparrow$217\%) \\
\end{tabular}
\label{table1}
\end{table*}

This section explores the effects of biased models in the performance of face recognition algorithms. Table \ref{table1} shows the performances obtained for each demographic group present in DiveFace.  Traditional face recognition benchmarks usually do not explore this kind of demographic covariates. Results reported in Table \ref{table1} exhibit large gaps between performances obtained by different demographic groups, suggesting that both gender and ethnicity significantly affect the performance of biased models. These effects are particularly high for ethnicity, with a very large degradation of the results for the class less represented in the training data (\textit{Group} 3 \textit{Female}). This degradation produces a relative increment of the Equal Error Rate (EER) of 196\% and 217\% for VGG-Face and ResNet-50, respectively, with regard to the best class (\textit{Group} 1 \textit{Male}). These differences are important as they mark the percentage of faces successfully matched and faces incorrectly matched. These results suggest that your ethnic origin can highly affect your possibilities to be incorrectly matched (false positives).

\subsection{Understanding biased performances}
\begin{figure}
\includegraphics[width=85mm]{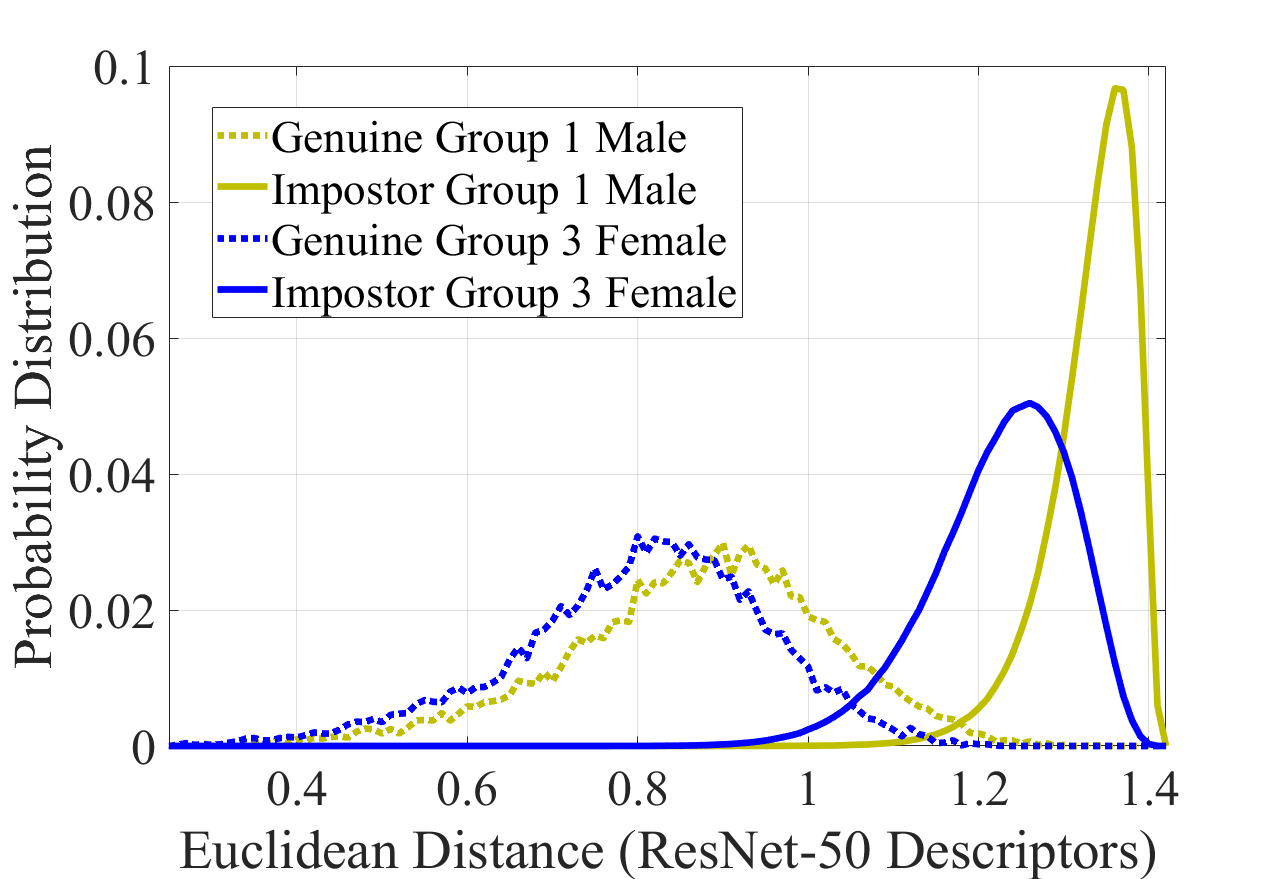} 
\caption{ResNet-50 face recognition score distributions for Group 3 females and Group 1 males.}
\label{Figure2}
\end{figure}

The relatively low performance in \textit{Group} 3 seems to be originated by a limited ability to capture the best discriminant features for the groups underrepresented in the training databases. The results suggest that features capable of reaching high accuracy for a specific demographic group may be less competitive in others. Let’s analyze the causes behind these degradations. Fig. \ref{Figure2} represents the probability distributions of genuine and impostor scores for \textit{Group} 1 \textit{Male} (the best group) and \textit{Group 3 Female} (the worst group). The comparison between genuine and impostor distributions reveals large differences for the impostor's ones. The genuine distribution (intra-class variability) between \textit{Group} 3 and \textit{Group} 1 is similar, but the impostor distribution (inter-class variability) is significantly different. The model has difficulties to differentiate face attributes from different subjects.

\textbf{Algorithmic discrimination implications:} define the performance function $f$ as the accuracy of the face recognition model, and $\textit{G}(\mathcal{D}_d^k)=f(\mathcal{D}_d^k,\theta^* )$ the goodness considering all the samples corresponding to class $k$ of the demographic criterion $d$, for an algorithm $\theta^*$ trained on the full set of data $\mathcal{D}$ (as described in Eq. \ref{eqn:goodnes_criterion}). Results suggest large differences between the goodness $\textit{G}(\mathcal{D}_d^k)$ for different classes, especially for classes $k=\textit{Group} \, 1,\textit{Group} \, 2,\textit{Group} \, 3$.

\subsection{Bias in face databases}\label{Bias in face databases}

\begin{figure}[t]
\includegraphics{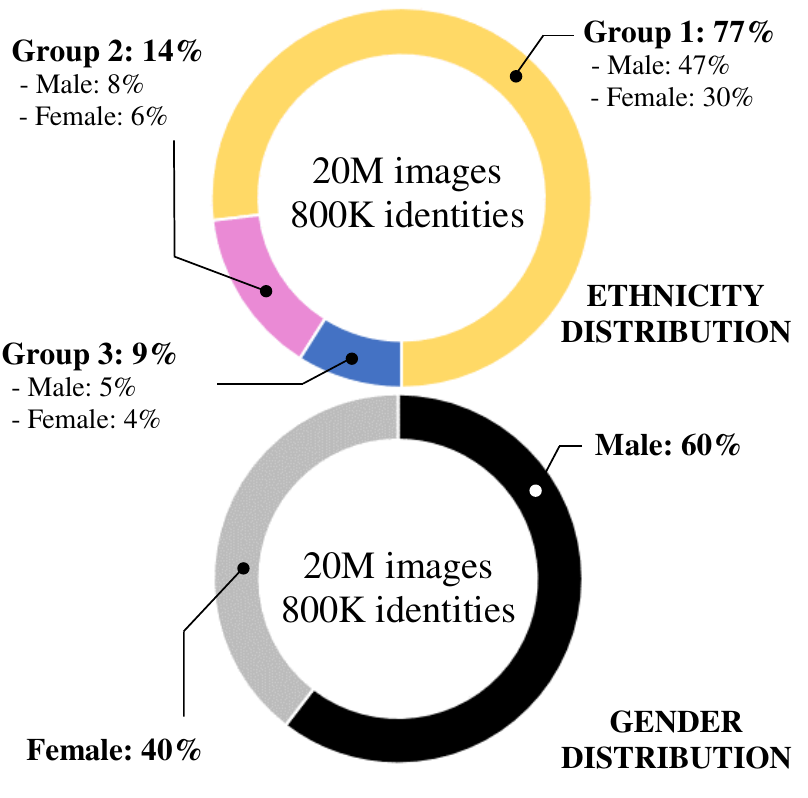} 
\caption{Demographic statistics of the 12 most cited face databases available in the literature. BioSecure \cite{21}, YouTubeFaces \cite{14}, PubFig \cite{17}, CasiaFace \cite{15}, VGGFace \cite{13}, CelebA \cite{16}, MS-Celeb-1M \cite{10}, Megaface \cite{11}, LFW \cite{20}, UTKface \cite{19}, VGGFace2 \cite{12}, IJB-C \cite{18}, DiveFace \cite{9}.}
\label{Figure3}
\end{figure}

Bias and discrimination concepts are related to each other, but they are not necessarily the same thing. Bias is traditionally associated with unequal representation of classes in a dataset. The history of automatic face recognition has been linked to the history of the databases used for algorithm training during the last two decades. The number of publicly available databases is high, and they allow training models using millions of face images. Fig. \ref{Figure3} summarizes the demographic statistics of some of the most cited face databases. Each of these databases is characterized by its own biases (e.g. image quality, pose, backgrounds, and aging). In this work, we highlight the unequal representation of demographic information in very popular face recognition databases. As it can be seen, the differences between ethnic groups are severe. Even though the people in \textit{Group} 3 are more than 35\% of the world's population, they represent only 9\% of the users in those popular face recognition databases.

Biased databases imply a double penalty for underrepresented classes. On the one hand, models are trained according to non-representative diversity. On the other hand, benchmark accuracies are reported over privileged classes and overestimate the real performance over a diverse society.

Recently, diverse and discrimination-aware databases have been proposed in \cite{3,25,wang2019mitigate}. These databases are valuable resources to explore how diversity can be used to improve face biometrics. However, some of these databases do not include identities \cite{3,25}, and face images cannot be matched to other images. Therefore, these databases do not allow to properly train or test face recognition algorithms.

\textbf{Algorithmic discrimination implications}: classes $k$ are unequally represented in the most popular face databases $\mathcal{D}$.

\subsection{Biased embedding space of deep models}

\begin{figure}[t]
\includegraphics{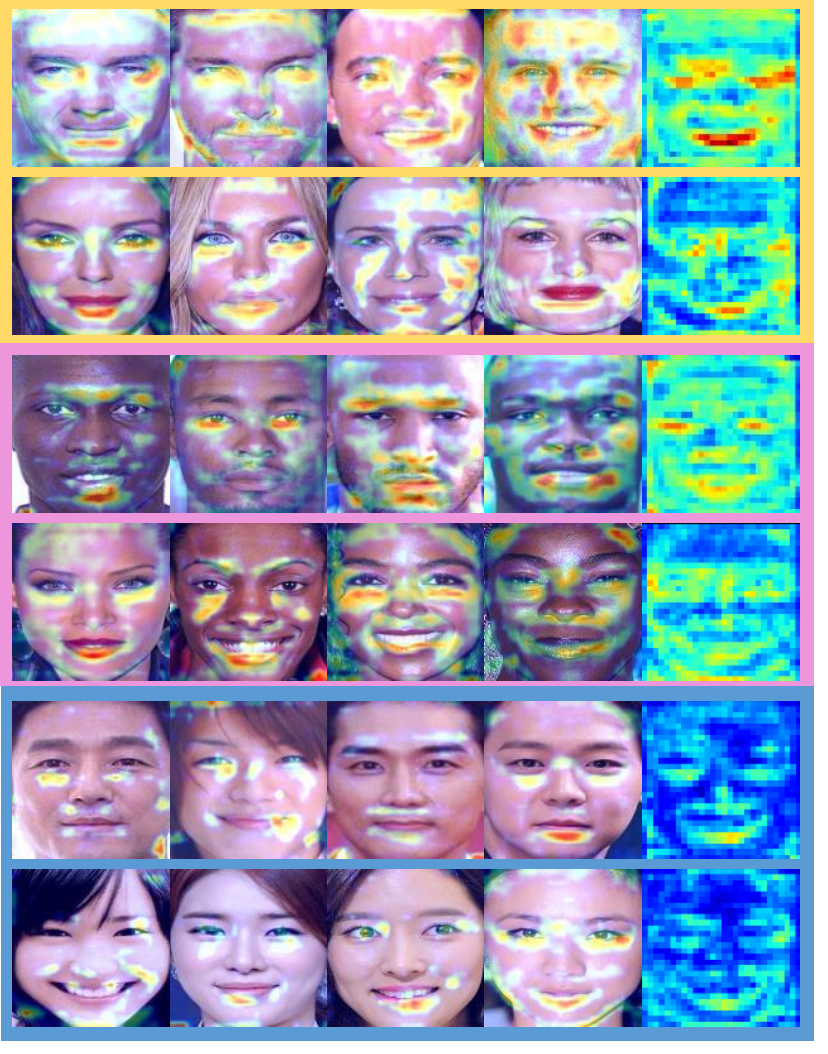} 
\caption{Examples of the six classes available in the DiveFace database (columns 1 to 4).  Column 5 shows the averaged Class Activation MAP (first filter of the third convolutional block of ResNet-50) obtained from 20 random face images from each of the classes. Columns 1-4 show Class Activation MAPs for each of the face images. Maximum and minimum activations are represented by red and blue colors respectively. Average pixel value of the activation maps generated for the six classes (\textit{Groups} 1 to 3, and \textit{Male}/\textit{Female}): G1M=0.23, G1F=0.19, G2M=0.21, G2F=0.18, G3M=0.12, G3F=0.13. (This is a colored image, see the digital version for a better quality.)}
\label{Figure4}
\end{figure}

We now analyze the effects of ethnicity and gender attributes in the embedding space generated by VGG-Face and ResNet-50 models. CNNs are composed of a large number of stacked filters. These filters are trained to extract the richest information for a pre-defined task (e.g. face recognition). As face recognition models are trained to identify individuals, it is reasonable to think that the response of the models can slightly vary from one person to another. In order to visualize the response of the model to different faces, we consider the specific Class Activation MAP (CAM) proposed in \cite{30}, named Grad-CAM. This visualization technique uses the gradients of any target concept, flowing into the selected convolutional layer to produce a coarse localization map. The resulting heat map highlights the activated regions in the image for the mentioned target (e.g. an individual identity in our case). Fig. \ref{Figure4} represents the heat maps obtained by the ResNet-50 model for faces from different demographic groups. Additionally, we include the heat map obtained after averaging results from 120 different individuals from the six demographic groups included in DiveFace. The activation maps show clear differences between ethnic groups with the highest activation for \textit{Group} 1 and the lowest for \textit{Group} 3. These differences suggest that features extracted by the model are, at least, partially affected by the ethnic attributes.

\begin{figure}[t]
\includegraphics[width=80mm]{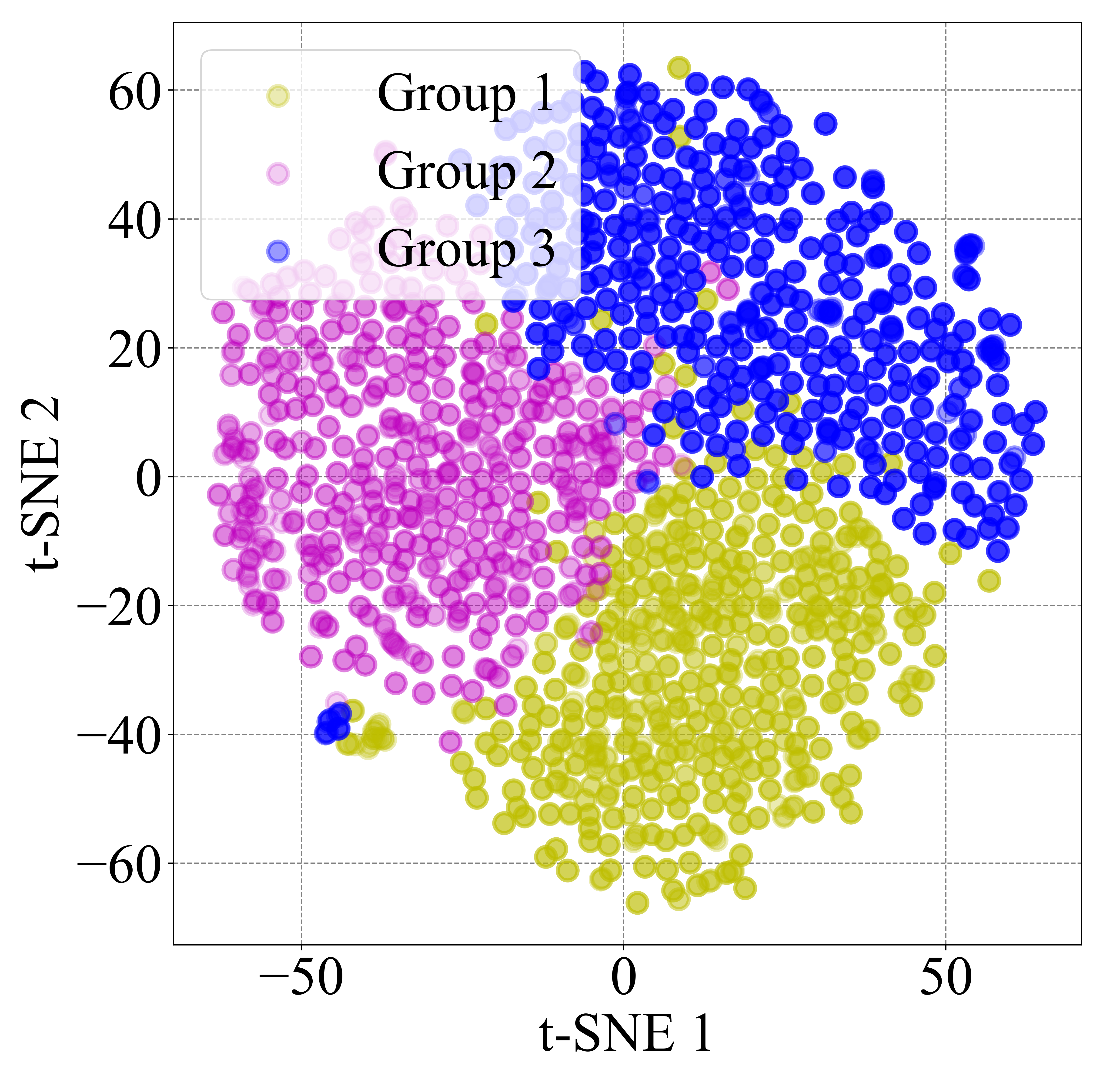} 
\caption{Projections of the ResNet-50 embeddings into the 2D space generated with t-SNE.}
\label{Figure5}
\end{figure}

On a different front, we applied a popular data visualization algorithm to better understand the importance of ethnic features in the embedding space generated by deep models. t-SNE is an algorithm to visualize high-dimensional data. This algorithm minimizes the Kullback-Leibler divergence between the joint probabilities of the low-dimensional embedding and the high-dimensional data. Fig. \ref{Figure5} shows the projection of each face into a 2D space generated from ResNet-50 embeddings and the t-SNE algorithm. Additionally, we have colored each point according to its ethnic attribute. As we can see, the resulting face representation results in three clusters highly correlated with the ethnicity attributes. Note that ResNet-50 has been trained for face recognition, not ethnicity detection. However, the ethnicity information is highly embedded in the feature space and a simple t-SNE algorithm reveals the presence of this information.

These two simple experiments illustrate the presence and importance of ethnic attributes in the feature space generated by face deep models.

\textbf{Algorithmic discrimination implications}: popular deep models trained for task \textit{T} on biased databases (i.e., unequally represented classes $k$ for a given demographic criterion $d$ such as gender) result in feature spaces (corresponding to the solution $\theta^*$ of the Eq. \ref{eqn:goodnes_criterion}) that introduce strong differentiation between classes $k$. This differentiation affects the representation $\textbf{x}$ and enables classifying between classes $k$ using $\textbf{x}$, even though \textbf{x} was trained for solving a different task ${T}$.

\section{Conclusions}

This work has presented a comprehensive analysis of face recognition models according to a new discrimination-aware perspective. This work presents a new general formulation of algorithmic discrimination with application to face recognition. We have shown the high bias introduced when training the deep models with the most popular databases employed in the literature, and testing with the DiveFace dataset with well balanced data across demographic groups\footnote{Available at GitHub: https://github.com/BiDAlab/DiveFace}. We have evaluated two popular models according to the proposed formulation. Biased models based on competitive deep learning algorithms have been shown to be very sensitive to gender and ethnicity attributes. This sensitivity results in different feature representations and a large gap between performances depending on the ethnic origin. This gap between performances reached up to 200\% of relative error degradation between the best class (\textit{Group} 1 \textit{Male}) and the worst (\textit{Group} 3 \textit{Female}). These results suggest that false positives are 200\% more likely in \textit{Group} 3 \textit{Female} than in \textit{Group} 1 \textit{Male} for the models evaluated in this work. These results encourage training more diverse models and developing methods capable to deal with the differences inherent to demographic groups. Future work will go in line with this approach, as authors do in \cite{wang2019mitigate}.

\section*{Acknowledgments}

This work has been supported by projects: BIBECA (RTI2018-101248-B-I00 MINECO/FEDER), Bio-Guard (Ayudas Fundacion BBVA a Equipos de Investigacion Cientifica 2017).

\bibliographystyle{aaai}
\bibliography{AAAI}

\end{document}